\documentclass[10pt,twocolumn,letterpaper]{article}

\usepackage{wacv}              
\usepackage{xcolor}
\definecolor{lightblue}{RGB}{16,98,180}
\definecolor{lightpink}{RGB}{243,40,109}
\definecolor{lightgreen}{RGB}{0,220,0}
\AtBeginDocument{%
  \hypersetup{%
    linkcolor=lightpink,
    citecolor=lightgreen,
    urlcolor=lightpink,
    filecolor=lightblue
  }%
}

\definecolor{wacvblue}{rgb}{0.21,0.49,0.74}
\usepackage[pagebackref,breaklinks,colorlinks,allcolors=wacvblue]{hyperref}

\def\wacvPaperID{553} 
\def\confName{WACV}
\def\confYear{2026}

\title{Tables Guide Vision: Learning to See the Heart through Tabular Data}

\author{
\makebox[\textwidth][c]{%
Marta Hasny$^{1,2}$ \quad
Maxime Di Folco$^{1,2}$ \quad
Keno Bressem$^{4}$ \quad
Julia A. Schnabel$^{1,2,3}$%
}\\[0em] 
\makebox[\textwidth][c]{$^{1}$ School of Computation, Information and Technology, Technical University of Munich, Germany}\\[-0.1em]
\makebox[\textwidth][c]{$^{2}$ Institute of Machine Learning in Biomedical Imaging, Helmholtz Munich, Germany}\\[-0.1em]
\makebox[\textwidth][c]{$^{3}$ School of Biomedical Engineering and Imaging Sciences, King’s College London, UK}\\[-0.1em]
\makebox[\textwidth][c]{$^{4}$ TUM University Hospital, Technical University of Munich, Germany}
}

\begin{document}
\maketitle

\begin{abstract}
Contrastive learning methods in computer vision typically rely on augmented views of the same image or multimodal pretraining strategies that align paired modalities. However, these approaches often overlook semantic relationships between distinct instances, leading to false negatives when semantically similar samples are treated as negatives. This limitation is especially critical in medical imaging domains such as cardiology, where demographic and clinical attributes play a critical role in assessing disease risk and patient outcomes. We introduce a tabular-guided contrastive learning framework that leverages clinically relevant tabular data to identify patient-level similarities and construct more meaningful pairs, enabling semantically aligned representation learning without requiring joint embeddings across modalities. Additionally, we adapt the k-NN algorithm for zero-shot prediction to overcome the lack of zero-shot capability in unimodal representations. We demonstrate the strength of our methods using a large cohort of short-axis cardiac MR images and clinical attributes, where tabular data helps to more effectively distinguish between patient subgroups. Evaluation on downstream tasks, including fine-tuning, linear probing, and zero-shot prediction of cardiovascular artery diseases and cardiac phenotypes, shows that incorporating tabular data guidance yields stronger visual representations than conventional methods that rely solely on image augmentation or combined image-tabular embeddings. Further, we show that our method can generalize to natural images by evaluating it on a car advertisement dataset. Code is available at \href{https://github.com/marteczkah/tables_guide_vision}{\texttt{this link}}.

\end{abstract}

\section{Introduction}
\begin{figure}[]
\includegraphics[width=\columnwidth]{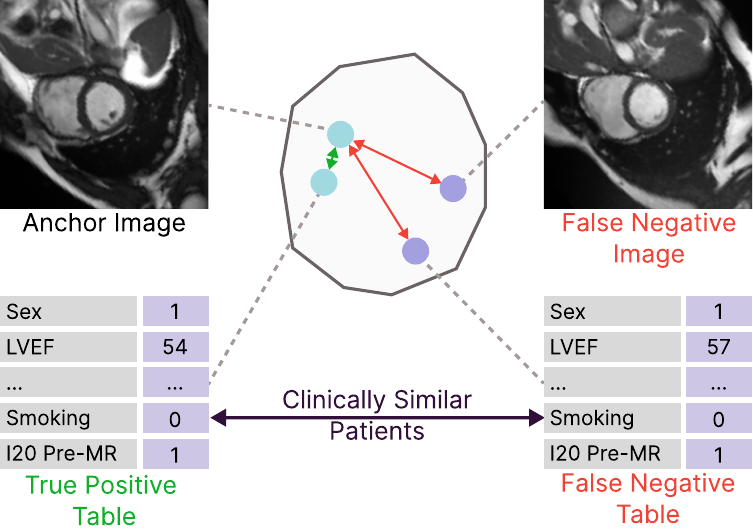}
\caption{Contrastive learning methods typically assume a one-to-one correspondence between positive pairs during training, an assumption present in both image augmentation and tabular supervision (similar to CLIP \cite{radford2021learning}, but with tabular data instead of text) approaches. This assumption can cause clinically similar patients (e.g., left and right side) to be treated as negative pairs, separating them in the embedding space and generating false negatives.}\label{fig_fn}
\vspace{-1.5em}
\end{figure}
Biobanks provide a vast quantity of multimodal medical data that can be leveraged to train medical foundation models. Typically, they include tabular data, such as demographic information and medical history, and imaging data, such as magnetic resonance (MR) or computed tomography (CT) scans. However, methods using images and tabular data still remain underexplored \cite{van2024tabular}. Meanwhile, in practice physicians often make clinical decisions based on both images and tabular attributes, as information such as patient demographics can provide crucial insights about the patient's health. Specifically in cardiology, attributes such as patient’s sex, age or smoking status are key factors in assessing the risk of cardiovascular diseases \cite{d2008general,arnett20192019}, which are the leading cause of death worldwide \cite{WHO2023}. Thus, developing methods that can integrate information from both images and tabular data is critical for improved clinical decision-making.  

Given this clinical challenge, contrastive learning has emerged as a powerful tool for multimodal data integration. In particular, multimodal contrastive learning using language supervision has been vastly explored as a way to train strong vision encoders \cite{zong2024self,bayoudh2022survey}. This approach was successfully extended to image–tabular data in \cite{hager2023best}, enabling the pretraining of image encoders supervised by tabular attributes and effectively leveraging the rich clinical information available in biobanks. However, the approach incorporated the standard rigid one-to-one correspondence between training samples used in contrastive learning, overlooking the semantic similarity between instances. As illustrated in Fig. \ref{fig_fn}, relying solely on image augmentations or image-tabular data pairs in contrastive learning can result in false negatives \cite{huynh2022boosting}, where clinically similar patients with the same diseases or phenotypes are incorrectly treated as dissimilar. This highlights the need for methods that can leverage tabular information to define clinically informed pair sampling strategies for contrastive learning. 

While biobanks provide structured tabular data ideal for training, such information is often unavailable in routine clinical practice, where time constraints and workflow pressures limit comprehensive data collection \cite{dugdale1999time}. Therefore, there is a need for methods that use tabular data during training to produce image encoders capable of accurate unimodal prediction at inference. Although multimodal training has gained popularity as a way to produce strong unimodal image encoders, it remains unclear whether such complexity is necessary. Recent studies show that unimodal training of visual representations can outperform language-based supervision in both natural images \cite{fan2025scaling} and medical settings \cite{perez2025exploring}. Motivated by those findings, we adapt a vision-centric approach, innovatively using tabular data as a proxy to guide the training of visual representations. 

To this end, our contributions are as follows:
\begin{itemize}
    \item We introduce \textbf{T}ables \textbf{G}uide \textbf{V}ision (TGV), a novel contrastive learning paradigm harnessing tabular data to construct clinically meaningful pair sampling strategy. Rather than enforcing a shared representation between modalities or relying on image augmentation, we use tabular similarity to construct multiple clinically-informed positive and negative pair assignments within images in a unimodal contrastive learning setting.
    \item We propose an approach to enable zero-shot prediction from unimodal visual representations by adapting the k-nearest neighbors (k-NN) algorithm. While k-NN aggregation is well-established in machine learning, it has not been systematically applied as an approach for unimodal zero-shot prediction in visual representations. We modify k-NN for this setting by using a representative set of labeled examples for each target class, against which query representations are compared. This representative set enables reliable mean label aggregation across neighbors, allowing for robust zero-shot prediction.
    \item We demonstrate the potential of TGV to leverage multimodal datasets for unimodal prediction by training a visual cardiac representation using MR images and tabular data from the UK Biobank \cite{sudlow2015uk} and evaluate it by predicting cardiac health outcomes and phenotype.
    \item We show that TGV generalizes beyond cardiology by conducting experiments on the DVM \cite{huang2023dvmcarlargescaleautomotivedataset} dataset, highlighting the potential of our method in other domains.
\end{itemize}

\section{Related Works}

\begin{figure*}[]
\includegraphics[width=\textwidth]{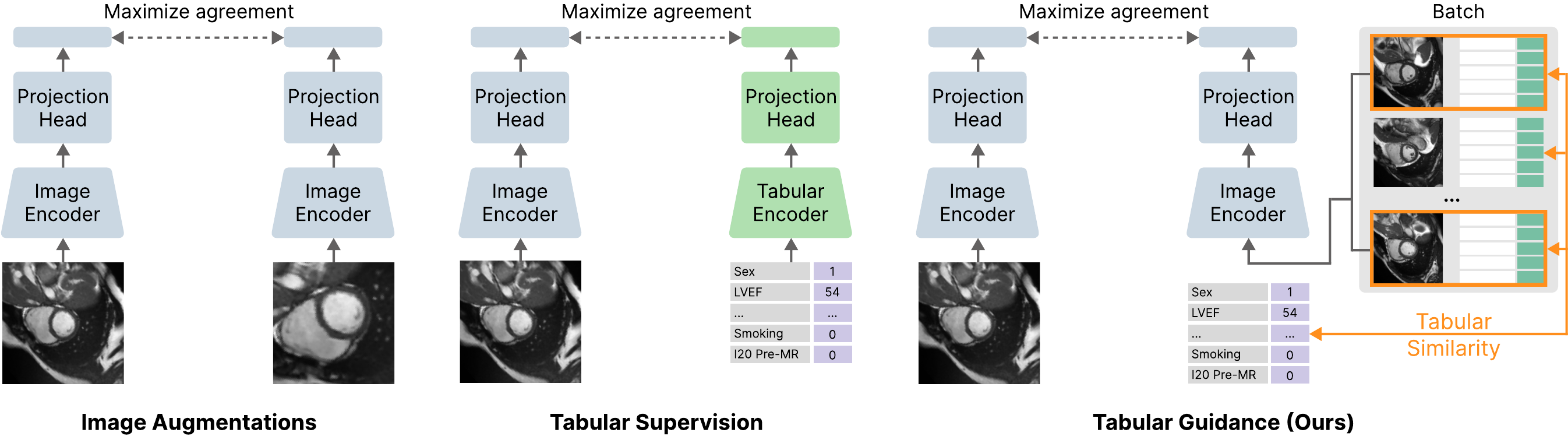}
\caption{Comparison of the proposed tabular guidance (TGV: Tables Guide Vision) approach against other contractive learning approaches. Instead of relying on different views of the same image for training, as in image augmentation approaches, TGV defines pairs between different subjects in the batch based on their tabular similarity. Tabular guidance operates without the need for a joint embedding space between images and tabular data, which is the case in tabular supervision.} \label{fig_method}
\vspace{-1em}
\end{figure*}

\subsection{Contrastive Learning}
Contrastive learning is a self-supervised representation learning approach that enables models to differentiate between similar and dissimilar samples by comparing pairs of instances. The objective is to align the representations of positive pairs, such as augmented views of the same image, while pushing apart the embeddings of negative pairs, which are typically other instances in the batch \cite{hadsell2006dimensionality}. By optimizing a contrastive loss function, such as InfoNCE \cite{oord2018representation}, the model learns an embedding space where semantically similar samples are clustered together, while dissimilar samples are separated. This paradigm has demonstrated strong performance in computer vision, giving rise to a range of influential methods, including SimCLR \cite{chen2020simple}, BYOL \cite{grill2020bootstrap}, and Barlow Twins \cite{zbontar2021barlow}, among others \cite{hu2024comprehensive}.

\subsection{Multimodal Contrastive Learning}

The success of contrastive methods in vision,  motivated their extension to multimodal data, facilitating alignment between text and images and enabling strong zero-shot performance. The first formulation of this paradigm was introduced in the medical domain with ConVIRT \cite{zhang2022contrastive}, which employs a dual-stream architecture composed of a separate image and text encoder. Using a bidirectional contrastive objective, ConVIRT learns a joint representations from paired medical image-text datasets. Subsequently, this approach was adapted and scaled to natural images in CLIP \cite{radford2021learning}, which utilizes large-scale web-scraped image-text pairs and demonstrated zero-shot results competitive with fully supervised models on standard benchmarks. Beyond image-text alignment, contrastive learning has also been extended to other combinations of modalities, such as image-genetics \cite{taleb2022contig, zhou2024mgi} or audio-video.\cite{jenni2023audiovisualcontrastivelearningtemporal,ma2020active}. 

\subsection{Multimodal Image-Tabular Learning}

Multimodal learning approaches has also been studied for image-tabular data. MMCL \cite{hager2023best} was the first to explore the concept of tabular data supervision in contrastive learning, leveraging rich tabular data to train a strong image encoder. The method uses images and tabular data for training, but is capable of unimodal predictions, similar to CLIP \cite{radford2021learning}. In contrast, subsequent work has largely focused on using both data modalities at inference. TIP \cite{du2024tip} introduced a self-supervised framework for image-tabular pretraining that is robust to missing or incomplete tabular data at inference. However, the method was not evaluated in a fully unimodal setting where tabular data is entirely absent during inference. STiL \cite{Du_2025_CVPR} proposed a semi-supervised, task-specific training framework that combines image and tabular modalities, but it relies on specialized training for each application and does not employ a general-purpose pretrained encoder capable of generalizing across multiple tasks. 

\subsection{Pairs in Contrastive Learning}

A key limitation of contrastive learning is that it overlooks semantic relationships between samples by assuming a rigid one-to-one correspondence between paired instances. As a result, semantically similar samples get pushed away from the anchor, leading to a problem of false negatives \cite{huynh2022boosting}. The issue is commonly addressed by redefining the pair assignments to account for inter-instance similarities. SoftCLIP \cite{gao2024softclip} adjusts the training objective by incorporating soft targets defined by fine-grained intra-modal similarity. FFF \cite{bulat2024fff} assigns multiple positive pairs by considering image-text, image-image, and text-text similarities. On the unimodal side, \cite{el2023learning} proposed using language similarity between text descriptions to sample semantically similar image pairs for contrastive visual representation learning.

The issue of false negatives has also been studied in the medical domain, where patients may exhibit clinical similarities, such as shared diseases. To address this, several works redefine their pair sampling strategies to incorporate for clinical similarities at training. MedCLIP \cite{wang2022medclip} uses clinical concepts information, while NB-CLIP \cite{wei2025relaxing} harnesses medical knowledge graphs to assign soft similarity scores between images and text. In a unimodal setting,  \cite{pai2024foundation} defines contrastive pairs in medical imaging by grouping 3D patches containing lesions as positives, while treating lesion-free regions as negatives. In \cite{holland2024metadata}, longitudinal patient metadata is used to guide pair selection for retinal OCT, though the approach relies on dataset-specific metadata and predefined rules, limiting its generalizability.

These prior works highlight the limitations posed by rigid pair definitions and the importance of incorporating semantic similarities into contrastive learning. While existing work has explored this challenge in vision-language approaches, the role of tabular data in guiding pair selection has remained largely unexamined. To the best of our knowledge, TGV is the first approach that can use large-scale tabular information to guide contrastive pair assignment between images of different subjects, providing a semantically-informed pairing strategy without introducing additional supervision or relying on handcrafted rules. 

\section{Methodology}

Tables Guide Vision leverages a multimodal dataset of image-tabular attributes pairs to learn a representation where clinically similar patients are aligned in the embedding space, guided by medical tabular data. Fig.~\ref{fig_method} shows how our method differs from image augmentation and tabular supervision based approaches. TGV does not rely on image augmentations, but rather defines the alternative views of the image as different, but clinically similar, subjects in the batch. In comparison with tabular supervision, our approach does not use a joint embedding space. The tabular data is used as a measure of similarity between the patients and is not used as input to any model.

\subsection{Vision Encoder}
TGV is based on one vision encoder, following the basic setting of SimCLR \cite{chen2020simple}. The vision encoder \(E\) encodes the images into embedding \(v \in \mathbb{R}^d\), followed by a projection head \(f_v\) mapping the embeddings to \(z \in \mathbb{R}^p\), where \(d\) and \(p\) are the dimensions of the embeddings. 
\begin{align}
      v = E(x) \\
      z = f_v(v)  ,
\end{align}
where \(x\) represents the input image. The projection head is only used for the training.

\subsection{Tabular Similarity}

In a batch of size \(N\), each image \(x_i\) is associated with a number of continuous and categorical attributes included in the tabular data \(a_i=\{a_{con_i}, a_{cat_i}\}\), which are used to calculate a similarity matrix \(S \in \mathbb{R}^{N \times N}\). As continuous and categorical features posses different mathematical properties their similarities are calculated separately and then combined into a single similarity matrix.

Categorical attributes from the batch are represented in a matrix \(a_{cat} \in \mathbb{R}^{N \times B}\), where \(B\) is the number of categorical attributes in the dataset after applying bipolar (-1/1) encoding. Categorical similarity \(S_{cat} \in R^{N \times N}\) is computed via cosine similarity using \(a_{cat}\).

Given a matrix of normalized continuous variables \(a_{con} \in \mathbb{R}^{N \times M}\), where \(M\) is the number of the attributes, we calculate a pairwise Euclidean distance matrix \(D \in \mathbb{R}^{N \times N}\). The distance is then converted to a similarity and normalized to match the range of categorical similarity, resulting in a continuous data similarity matrix \(S_{con}\). The continuous and categorical similarities are then used to calculate the combined similarity between the samples in the batch using:

\begin{equation}
\label{eq_attribute}
    S = \lambda S_{con} + (1-\lambda)S_{cat},
\end{equation}
where \(\lambda\) is a hyperparameter weighting the tabular guidance towards continuous or categorical attributes.

\subsection{Tabular Data-Guided Visual Learning}

The constructed tabular similarity matrix \(S\) is used to define the training image pairs based on intra-batch similarity. As multiple patients in the batch can be highly similar to each other, an image may have multiple corresponding pairs. For each image \(x_i\) in the batch, the most similar image \(x_j\) is the one with the highest corresponding similarity score:
\begin{equation}
    x_j = \arg\max_{j \neq i} S_{ij}, \; \text{for} \; i, j \in \{1, \dots, N\}.
\end{equation}
Additionally, a threshold \(h\) is introduced to allow for sampling of multiple positive pairs, as many subjects in the batch can resemble high tabular similarity. For each anchor image \( x_i \), let the maximum similarity score to any other image in the batch be:
\begin{equation}
    M_i = \max_{j \ne i} S_{ij}
\end{equation}
Then, the set of valid positive pairs for \( x_i \) can be defined as,
\begin{equation}
\label{eq_thresh}
   \mathcal{P}_i = \left\{ x_j \,\middle|\, j \ne i,\; S_{ij} \ge M_i - h \right\} 
\end{equation}
This approach ensures that all highly similar instances contribute to the contrastive learning process, leveraging intra-batch relationships. With all the pairs in the batch constructed, the contrastive loss function aligns the representations such that the paired ones are pulled closer to each other, while the unpaired ones are pulled away from one another. The loss is defined as:
\begin{equation}
    L = \frac{1}{N} \sum_{i=1}^{N} -\log \left( \frac{\sum_{j \in \mathcal{P}_i} \exp{(\langle z_i,z_j \rangle / \tau)}}{\sum_{j=1}^{N} \exp{(\langle z_i,z_j \rangle / \tau)} }\right),
\end{equation}

where \(\langle \cdot, \cdot \rangle\) stands for cosine similarity, \(\tau\) is a temperature parameter scaling the logits.

\subsection{Unimodal Zero-Shot Prediction}

The setting of our proposed unimodal zero-shot prediction is based on k-NN. It can be considered similar to the zero-shot approach presented in image-text joint representation spaces \cite{radford2021learning}. However, instead of computing similarities between images and text embeddings, we make predictions based on similarities between images in the learned embedding space. 

Given a representative sample of images from the training set, i.e., one that captures a range of values for the predicted attribute, we generate representations of those images as a reference set of potential pairings \(P = \{v_j | x_j \in \text{training set} \}\). To make the prediction for an unseen image \(x_i\), the cosine similarity is calculated between its embedding \(v_i\) and all embeddings \(v_j\) in \(P\), e.g.,
\begin{equation}
    s_{ij} = \frac{\langle v_i, v_j \rangle}{\| v_i \| \| v_j \|}, \quad \forall v_j \in P.
\end{equation}  
The prediction for a given tabular attribute \(a_i\) is then calculated as the mean of that attribute across the top \(K\) most similar images:
\begin{equation}
    \hat{a}_i = \frac{1}{K} \sum_{j \in \mathcal{N}_i} a_j, 
    \label{eq_knn}
\end{equation} 
where \( \mathcal{N}_i \) is the set of indices corresponding to the \( K \) most similar images based on \( s_{ij} \).  
\section{Experiments}

\begin{table*}
\centering
    \resizebox{\linewidth}{!}{%
\small
\begin{tabular}{lcccccccccccccc}
\toprule
     & \multicolumn{2}{c}{CAD \(\uparrow\)} & \multicolumn{2}{c}{LVEF \(\downarrow\)} & \multicolumn{2}{c}{LVEDM \(\downarrow\)} & \multicolumn{2}{c}{LVEDV \(\downarrow\)} & \multicolumn{2}{c}{RVEF \(\downarrow\)} & \multicolumn{2}{c}{RVEDV \(\downarrow\)} & \multicolumn{2}{c}{MYOESV \(\downarrow\)} \\
     \midrule 
   Model  & ZS         & FT         & ZS          & FT         & ZS          & FT          & ZS          & FT          & ZS          & FT         & ZS          & FT          & ZS           & FT          \\
     \midrule
Mean-Guess    & - & - & 4.81 & - & 17.85 & - & 29.54 & - & 4.73 & - & 26.73 & - & 17.85 & -  \\ \midrule
\textit{Supervised} &  &  &  &  &  &  &  &  &  &  &  &  &  &   \\
ResNet50 \cite{hara2018can} & - & 65.61 & - & 4.31 & - & 5.59 & - & 10.27 & - & 3.81 & - & 8.55 & - & 6.44  \\ \midrule
\textit{Image Augmentation} &  &  &  &  &  &  &  &  &  &  &  &  &  &   \\
SimCLR \cite{chen2020simple} & 62.05 & 71.68 & 4.72 & 3.49 & 13.01 & \underline{5.25} & 23.26 & \underline{9.82} & 4.71 & 3.18 & 21.89 & 8.25 & 13.48 & 6.06  \\
BYOL \cite{grill2020bootstrap} & 56.99 & 67.32 & 4.92 & 3.99 & 16.43 & 5.74 & 27.81 & 10.31 & 4.95 & 3.39 & 26.29 & 7.99 & 16.19 & 5.96  \\
SimSiam \cite{chen2021exploring}  & 57.01 & 69.89 & 4.93 & 3.93 & 16.04 & 6.59 & 27.46 & 11.09 & 4.94 & 3.43 & 25.62 & 8.52 & 16.14 &  5.89 \\
Barlow Twins \cite{zbontar2021barlow}  & 55.12 & 65.01 & 4.90 & 3.57 & 16.54 & 6.12 & 27.17 & 11.01 & 4.97 & 3.39 & 25.98 & 8.35 & 16.52 &  6.14 \\ \midrule
\textit{Tabular Supervision} &  &  &  &  &  &  &  &  &  &  &  &  &  &   \\ 
MMCL \cite{hager2023best}  & 62,49 & \underline{72.91} & \underline{4.48} & \underline{3.27} & \underline{8.74} & 5.63 & \underline{15.12} & 9.95 & \underline{4.55} & \underline{3.12} & \underline{13.70} & \underline{7.47} & \underline{9.54} & \underline{5.51}  \\ \midrule
\textit{Tabular Guidance} &  &  &  &  &  &  &  &  &  &  &  &  &  &   \\
TGV (Ours)& \textbf{68.70} & \textbf{76.1} & \textbf{4.08} & \textbf{3.18} & \textbf{7.64} & \textbf{4.86} & \textbf{13.63} & \textbf{9.23} & \textbf{3.98} & \textbf{2.95} & \textbf{12.43} & \textbf{7.39} & \textbf{8.18} & \textbf{5.2}  \\
\bottomrule
\end{tabular}
    }
    \caption{Downstream task performance comparison for multi-label CAD classification measured using AUC and cardiac phenotype prediction (remaining columns, \(\downarrow\)) using MAE. ZS stands for zero-shot, FT for fine-tuning. The best result is shown in \textbf{bold}, while the second-best is \underline{underlined}.}
\label{tab_compare}
\vspace{-1em}
\end{table*}

\subsection{Dataset}

We use the UK Biobank \cite{sudlow2015uk} population study for the training and evaluation of our method. The dataset contains 49,737 pairs of short-axis cardiac MR and tabular data, out of which 39,975 are used for training, 2,794 for validation, and 6,968 for testing. We use volumes of the cardiac MR as the input to our image encoder consisting of 11 slices over 10 frames, uniformly sampled from the initial 50 frames. The images are zero-padded to a square and then cropped to a size 128$\times$128. Our tabular data consists of 24 attributes, including 10 categorical and 14 continuous variables (see supplementary materials (SM) Table 1 for full list). The continuous attributes consist of the cardiac phenotype, e.g., left and right ventricular ejection fraction. The categorical data consist of demographics information, including sex and smoking, and attributes on the presence of CAD disorders split into those diagnosed pre- and post-MR. The attributes include four different CAD (angina pectoris, acute myocardial infarction, other acute ischemic heart disease, and chronic ischemic heart disease) following ICD-10 definition as described in \cite{hager2023best}. Due to the low frequency of CAD (8\% of the dataset), we construct a disease-balanced training sample by including all CAD-positive patients and an equal number of healthy controls, yielding 6,426 training samples for multilabel CAD prediction. We only consider diseases reported before the scan date for the classification task. For the fine-tuning of the cardiac phenotype predictions, we use a sample of 5,000 images, with the label quality check performed as described in \cite{bai2018automated}. Further, three non-overlapping, disease-balanced samples of 2,000 patients are considered as the representative set \(P\), used for the zero-shot prediction tasks. The zero-shot predictions are based on at most 20\% of the most similar embeddings in the reference set for CAD classification and 2.5\% for cardiac phenotype prediction. The final percentage is defined by empirical tuning on the validation set (see SM Sec. C.3). 

\subsection{Implementation} 
\label{sec_implement}

All image encoders are 3D ResNet50 models implemented using the MONAI framework \cite{cardoso2022monai}. The embedding dimensions are set to 2048 for the encoder and 128 for the projection head, following SimCLR \cite{chen2020simple}. The similarity threshold \(h\) is set to 0.05, the regularization parameter \(\lambda\) to 0.5, and the temperature parameter \(\tau\) to 0.1. The models are all pretrained for 10 epochs to ensure fair comparison. We use batch size of 512 and learning rate of 1e-3. For downstream tasks, the projection head is removed, and a linear layer is added on top of the image encoder, with an output size corresponding to the number of classes. All downstream tasks are fine-tuned for 35 epochs. Baseline pretraining follows the authors' original settings \cite{hager2023best, chen2020simple, grill2020bootstrap, zbontar2021barlow, chen2021incremental}, modified to match our 10-epoch schedule. See SM Sec. B for details. The supervised models are trained on the training sets prepared for the fine-tuning.
\section{Results}

\begin{figure*}[]
    \centering
    \includegraphics[width=\linewidth]{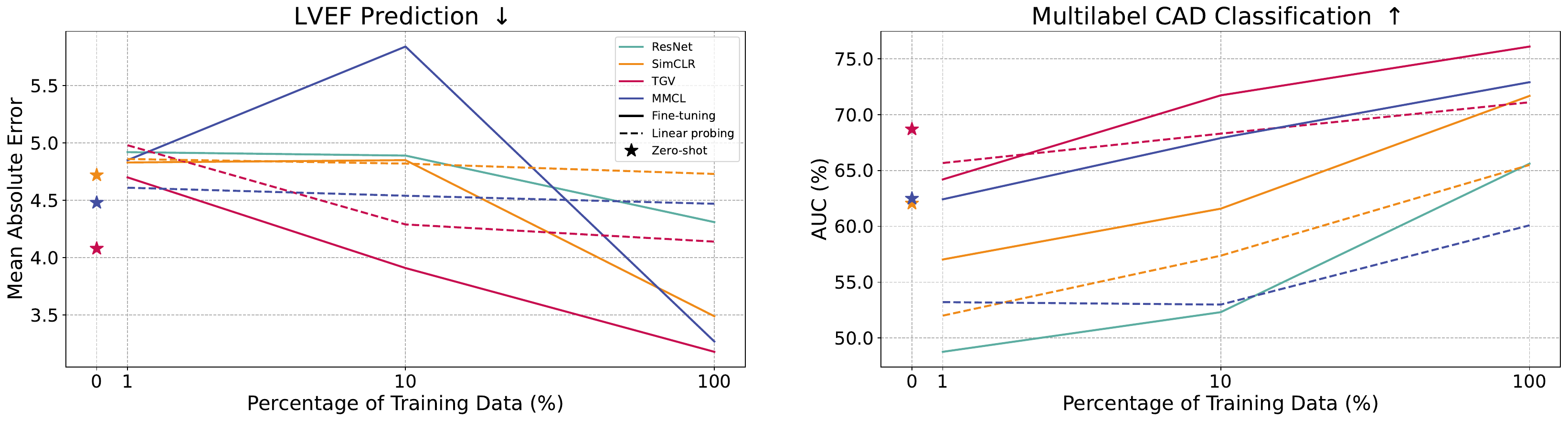}
    \caption{Performance of the models under different amount of training data on LVEF prediction (left, lower is better) and multilabel CAD classification (right, higher is better) tasks. The performance is evaluated using fine-tuning (solid lines), linear probing (dashed lines), and zero-shot prediction (stars). }
    \label{fig_limited}
    \vspace{-1em}
    
\end{figure*}

\subsection{Tabular Guidance Outperforms Image Augmentation \& Tabular Supervision}
We compare the performance of our method against a supervised ResNet50 \cite{he2016deep} and five contrastive learning baselines: four image-only methods, SimCLR \cite{chen2020simple}, SimSiam \cite{chen2021exploring}, Barlow Twins \cite{zbontar2021barlow}, and BYOL \cite{grill2020bootstrap}, and one image-tabular method, MMCL \cite{hager2023best}, which uses tabular supervision (as illustrated in Fig. \ref{fig_method}). All methods are evaluated across two downstream task categories. First, we report the performance on multilabel CAD classification using the area under the receiver operating characteristic curve (AUC), as it provides a robust evaluation metric for datasets with a low frequency of non-healthy samples. Second, we evaluate cardiac phenotype prediction using mean absolute error (MAE) across six attributes: left ventricular ejection fraction (LVEF), left ventricular end-diastolic mass (LVEDM), left ventricular end-diastolic volume (LVEDV), right ventricular ejection fraction (RVEF), right ventricular end-diastolic volume (RVEDV), and myocardial end-systolic volume (MYOESV). These attributes are selected to comprehensively assess the models' performance across two characteristic phases of the cardiac cycle and to ensure coverage of the three main anatomical regions of the heart. Table \ref{tab_compare} reports the results for both zero-shot prediction and fine-tuning. The zero-shot results are reported as a mean over the three representative sets \(P\). To evaluate the robustness of the zero-shot predictions, we report the standard deviations over the three sets in SM Table 4, while SM Table 3 reports results over different sizes of the three sets.  

TGV achieves the best performance across all the tasks on both fine-tuning and zero-shot predictions, demonstrating the strength of tabular data guidance in training visual representations. This shows that defining clinically meaningful pairs yields stronger representations than image augmentation-based pair sampling, further showing the importance of incorporating the semantic information into the pair sampling process. Additionally, tabular guidance also outperforms tabular supervision, suggesting that training a joint embedding could limit the encoder's ability to learn directly from the images themselves due to a strong supervisory signal being provided by the other modality. In contrast, tabular guidance directs the model to learn clinically meaningful information embedded in the tabular attributes directly from the images, resulting in a stronger visual encoder. Furtermore, to show that TGV can generalize to other datasets and domains, we provide experiments on the DVM \cite{huang2023dvmcarlargescaleautomotivedataset} dataset in the SM Section D.

\subsection{TGV is Effective under Limited Data}
Data annotation can be costly and time-consuming, especially in the medical domain. To evaluate the efficiency of the models in low-data regime, we sample 1\% (CAD: 64, LVEF: 50 samples) and 10\% (CAD: 642, LVEF: 500 samples) of the original classification and phenotype prediction training sets and use it for fine-tuning and linear probing. The validation and test sets are unchanged. Fig. \ref{fig_limited} presents the results of the models trained with limited data. For clarity, we only include MMCL \cite{hager2023best} and SimCLR \cite{chen2020simple}, while the results of the other baselines are presented in the SM Fig. 1. TGV achieves the best performance across all data regimes for both CAD classification and LVEF prediction. At the 10\% data regime, TGV matches or outperforms SimCLR and MMCL trained on the full dataset for CAD classification, under both fine-tuning and linear probing. These findings highlight the strength of the representation learned by our method, demonstrating its utility in settings with limited data, such as rare diseases.

\subsection{k-NN Allows Unimodal Zero-Shot Prediction}
\begin{figure*}[]
    \centering
    \includegraphics[width=\textwidth]{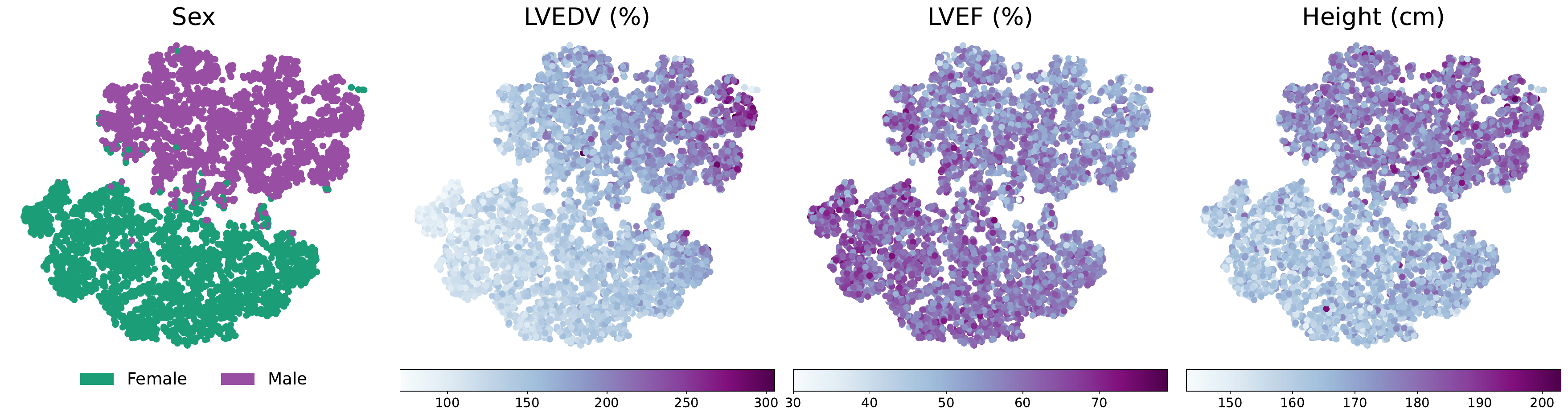}
    \caption{t-SNE visualization of the feature embedding generated with TGV. Sex, LVEF, and LVEDV have been included as attributes for calculation of tabular similarity during training. Height was not included.}
    \label{fig_tsne}
    \vspace{-1em} 
\end{figure*}

The lack of zero-shot abilities is one of the crucial limitations of unimodal representations. We evaluate the performance of the unimodal zero-shot approach in comparison with fine-tuning and linear probing. While fine-tuning on the full dataset achieves the highest overall performance, our zero-shot method often surpasses fine-tuning when labeled data is limited, and consistently outperforms linear probing across all data regimes for LVEF prediction (Fig. \ref{fig_limited}). This strong LVEF prediction performance is likely because LVEF is a continuous and widely available measure, enabling well-clustered representations (Fig. \ref{fig_tsne}) that support accurate zero-shot regression. Although CAD classification is more challenging due to class imbalance (8\% positive cases), zero-shot performance stands competitive on this task in low-data regimes.

\subsection{TGV Builds Clinically Aware Embeddings}
\noindent \textbf{Does TGV capture clinically meaningful structure in its latent space?} To assess the quality of the learned representations, we visualize the t-SNE of test set embeddings, colored by sex, height, LVEF, and LVEDV. The results for TGV are shown in Fig.~\ref{fig_tsne} (see SM Fig. 3 for SimCLR \cite{chen2020simple} and MMCL \cite{hager2023best}). TGV is able to generate representations where similar phenotypes are clustered at close locations. Furthermore, height is not included in the tabular similarity calculations at training, yet we can still see proper clustering of these attribute. The generated embeddings exhibit a primary separation based on sex, which is consistent with established cardiac physiological differences between sexes \cite{st2022sex}. These observations suggest that TGV captures underlying phenotype structure in a clinically meaningful way, even beyond explicitly provided training signals. 

\begin{table}[]
    \centering
        \caption{Zero-shot demographics prediction. Sex is measured using accuracy, while the other attributes using MAE. The * corresponds to attributes that were not included at pretraining.}
    \resizebox{\columnwidth}{!}{%

\label{tab_demo}
\centering
\begin{tabular}{lcccc}
\toprule
 & Sex $\uparrow$ &  BMI* $\downarrow$  & Height* $\downarrow$ & Weight* $\downarrow$  \\ \midrule
Mean-Guess & - & 3.31 & 7.64 & 11.78          \\   
SimCLR \cite{chen2020simple} & 75.04 & \underline{2.54} & 6.24 & \underline{8.43}            \\
BYOL \cite{grill2020bootstrap} & 69.72 & 3.33 & 7.02 & 10.44            \\
SimSiam \cite{chen2021exploring} & 68.19 & 2.86 & 6.93 & 8.80           \\
Barlow Twins \cite{zbontar2021barlow} & 72.90 &  3.16 & 7.02 & 10.88          \\
MMCL \cite{hager2023best} & \underline{92.60} & 2.98 & \underline{5.12} & 8.82            \\
TGV (Ours) & \textbf{98.16} & \textbf{2.39} & \textbf{4.88} & \textbf{7.44}            \\ \bottomrule
\end{tabular}

    }
    \label{tab_demo}
    \vspace{-1em}
\end{table}

\noindent \textbf{Can tabular guidance learn demographics?} Clinicians commonly use demographic information in their decision making. Thus, it is crucial for an image encoder to capture that information. We perform a zero-shot prediction of four demographic attributes, including sex, weight, height, and body-mass index (BMI). Sex prediction is performed as classification and measured using accuracy, while the other metrics are predicted as a value and evaluated using MAE. Only the sex attribute was included during training, meaning no information about weight, height, or BMI was incorporated to guide the training process. As shown in Table \ref{tab_demo}, TGV achieved the best performance at capturing the demographics information in its representation space. This suggests that tabular-data guidance allows the encoder to capture information encoded in the image, such as weight or height, even without explicitly incorporating those attributes at training. MMCL \cite{hager2023best} outperforms image augmentation methods in sex and height prediction, but fails to image augmentation-based SimCLR \cite{chen2020simple} on BMI and weight. This result further solidifies our claim that tabular data supervision might limit the encoder's ability to capture information directly from the images by providing a strong supervisory signal in a form of a different modality, reducing the strength of the image encoder itself. 

\subsection{Ablation Studies}

\begin{table}[]
    \centering
        \caption{Performance under different values of threshold\textit{ h} (Eq. \ref{eq_thresh}), which controls the multi-pair approach.}
    \small
    \label{tab_thresh}
\centering
\small
\begin{tabular}{lcccc}
\toprule
Threshold \textit{h} & \multicolumn{2}{c}{CAD \(\uparrow\)} &  \multicolumn{2}{c}{LVEF \(\downarrow\)} \\ \midrule
  & ZS & FT & ZS & FT       \\ \midrule
0.0  & 56.35 & 72.06 & 4.90 &   3.89       \\
0.05 & \textbf{68.70} & \textbf{76.1} & \textbf{4.08} & \textbf{3.18} \\
0.1  & 67.17 & 74.41 & 4.37 &  3.47 \\
0.2  & 67.60 &  74.63 & 4.38 & 3.33 \\ \bottomrule
\end{tabular}


    \label{tab_thresh}
    \vspace{-1em}
\end{table}

\begin{table*}[]
\centering
\caption{Impact of parameter \(\lambda\) (Eq.~\ref{eq_attribute}) on the transferability of TGV for downstream tasks using fine-tuning (FT) and zero-shot (ZS) predictions. \(\lambda>0.5\) emphasizes continuous attributes (Con), while \(\lambda<0.5\) favors categorical fields (Cat) in similarity computations.}
\small
\begin{tabular}{lccccccccccccccc}
\toprule
    & & \multicolumn{2}{c}{CAD \(\uparrow\)} & \multicolumn{2}{c}{LVEF \(\downarrow\)} & \multicolumn{2}{c}{LVEDM \(\downarrow\)} & \multicolumn{2}{c}{LVEDV \(\downarrow\)} & \multicolumn{2}{c}{RVEF \(\downarrow\)} & \multicolumn{2}{c}{RVEDV \(\downarrow\)} & \multicolumn{2}{c}{MYOESV \(\downarrow\)} \\
     \midrule 
  & \(\lambda\)  & ZS         & FT         & ZS          & FT         & ZS          & FT          & ZS          & FT          & ZS          & FT         & ZS          & FT          & ZS           & FT          \\
     \midrule
\textit{Con} & 1    & 62.0 & \underline{74.05} & \underline{4.36} & \underline{3.23} & \underline{7.68} & \underline{5.17} & \textbf{12.94} & \underline{9.57} & \underline{4.24} & \underline{3.07} & \textbf{11.47} & \underline{7.55} & \underline{8.17} & \underline{5.39}  \\
& 0.75 & 64.39 & 73.83 & \underline{4.36} & 3.32 & 8.82 & 5.47 & 14.22 & 9.74 & 4.4 & 3.16 & \underline{12.16} & 7.65 & 9.35 & 5.59  \\
& 0.5 & \textbf{67.97} & \textbf{76.1} & \textbf{4.08} & \textbf{3.18} & \textbf{7.65} & \textbf{4.86} & \underline{13.7} & \textbf{9.23} & \textbf{3.98} & \textbf{2.95} & 12.38 & \textbf{7.39} & \textbf{8.16} & \textbf{5.2}  \\
& 0.25 & 65.32 & 71.39 & 4.52 & 3.39 & 12.06 & 5.89 & 19.95 & 10.28 & 4.57 & 3.3 & 18.76 & 7.92 & 12.29 & 6.19  \\
\textit{Cat} & 0  & \underline{65.44} & 71.38 & 4.55 & 3.49 & 11.18 & 5.98 & 21.46 & 10.62 & 4.6 & 3.37 & 20.54 & 8.11 & 11.4 & 6.11  \\
\bottomrule
\end{tabular}

\label{tab_attr}
\vspace{-1em}
\end{table*}
\noindent \textbf{One-to-one or many-to-many correspondences in contrastive learning?} Assuming a perfect one-to-one correspondence often hinders the performance of contrastive learning. We evaluate the effectiveness of our multi-pair approach introduced via a hyperparameter \(h\) (Eq. \ref{eq_thresh}) by performing experiments on the effect of different threshold values on the downstream performance of CAD classification and LVEF prediction. At \(h=0\) each anchor is only paired with one instance, resulting in a single pair approach. Increasing the value of \(h\) results in more positive pairs per anchor. As presented in Table \ref{tab_thresh}, TGV achieves the best performance at \(h=0.05\). The single-pair approach (\(h=0\)), achieves the worst performance, showing the need for many-to-many approach, rather than a simple one-to-one correspondence. However, setting the value of \(h\) too high could result in too many pairings, causing the model to represent overly broad or noisy relationships between samples. This can lead to less discriminative embeddings, as the model may struggle to maintain sufficient uniformity in the representation space and risk collapsing semantic distinctions between different samples. In extreme cases, this could result in a form of representation collapse, where embeddings for diverse samples become undesirably similar, undermining the effectiveness of the learned representations. On the other hand, setting the parameter too low could miss on semantic similarity between the samples and result in false negatives. The value of \(h\) should be tuned based on a given dataset and its underlying relationships between samples.

\noindent \textbf{Which tabular attributes contribute most to the embedding quality?} To evaluate which features are of most importance for tabular guidance, we evaluate the impact of weighting tabular guidance towards categorical or continuous data by experimenting with different values of the hyperparameter \(\lambda\) from Eq.~\ref{eq_attribute}. The results are presented in Table~\ref{tab_attr}. Values of \(\lambda\) below 0.5 increase the influence of categorical features, while \(\lambda\) above 0.5 puts stronger emphasis on continuous data. Continuous attributes mostly consist of cardiac phenotype features and describe the function and structure of the heart. Many of them could be described as morphological information. Categorical data mostly describes the presence of cardiovascular disorders and demographic information. 

The model trained with \(\lambda=0.5\) achieved the best results, outperforming the other models in nearly all zero-shot and all fine-tuning tasks, suggesting that the best performance is achieved when categorical and continuous attributes contribute to the guidance in the same amount. Strong influence of categorical attributes (\(\lambda < 0.5\)) yields the poorest results. As our dataset has a very small frequency of samples with diseases (8\%) and the categorical attributes consist mostly of those describing the presence of CAD, there might not be enough variability between the samples to properly guide the training.  Increased influence of continuous attributes (\(\lambda > 0.5\)) lowers the performance in comparison to equal guidance, but not as much as strong categorical guidance. This suggests that while continuous attributes retain variability across samples, their differences may not provide sufficiently descriptive signals for tabular guidance. Furthermore, a better performance under strong continuous guidance is consistent with Hager at al. \cite{hager2023best}, where morphological features improved embeddings. The parameter \(\lambda\) should be adjusted per dataset, taking into consideration the distribution of its tabular attributes.

\section{Limitations}
Our work proposes adapting k-NN for unimodal zero-shot prediction. However, unlike image-text zero-shot methods, our approach requires access to a representative sample of images for the target task. Acquiring such a sample may be particularly challenging in the context of rare diseases, where data availability is limited. Future work could address this limitation by generating synthetic samples of rare diseases using generative models.

Another important limitation is the demographic imbalance in the UK Biobank dataset, which predominantly consists of white subjects. Individuals from other ethnic backgrounds comprise only around 5\% of the dataset. This imbalance may lead to biased predictions for underrepresented groups, particularly given known disparities in cardiac disease risk across different ethnicities \cite{lopez2022racial}. Ensuring fair model performance across diverse populations remains a critical area for future research.
\section{Conclusion}

We present TGV: Tables Guide Vision, a contrastive learning paradigm leveraging tabular data to generate clinically meaningful pairs for training of visual representations. Our approach outperforms image augmentation-based contrastive learning and tabular supervision on coronary artery disease classification and cardiac phenotype prediction, highlighting the strength of our approach in a medical setting. Additionally, we adapt k-NN to serve as a robust zero-shot prediction approach in unimodal image representations, overcoming a crucial limitation of this setting. Through zero-shot experiments, we show that our method outperforms previous approaches at capturing demographic information in its representation space. TGV can be leveraged to train medical foundation models grounded on rich clinical information, paving the way for more robust and generalizable medical models. While we designed the method with a clinical setting in mind, experiments on natural images show that the approach is generic and can generalize to image-tabular datasets in other domains without incorporating any changes.
\section*{Acknowledgments}
This research has been conducted using the UK Biobank Resource under Application Number 87065. MH is in part supported by the Munich School of Data Science (MUDS) and the European Laboratory for Learning and Intelligent Systems (ELLIS) PhD program.

{
    \small
    \bibliographystyle{ieeenat_fullname}
    \bibliography{main}
}
\newpage
\appendix
\section{Detailed Data Description}

\subsection{Tabular Attributes}

Table \ref{tab_attrlist} presents a comprehensive list of tabular attributes from the UK Biobank that were used for tabular similarity calculation during pretraining. These attributes were consistently used across all baseline methods that incorporated tabular data during pretraining. Attributes marked as \textit{extracted} were derived from automated cardiovascular MRI analysis, as described in \cite{bai2018automated}. Additionally, Table \ref{tab_attrother} lists attributes that were used for evaluation but were not included in the pretraining process, along with their corresponding field IDs.

\subsection{Multilabel CAD Classification Details}

One of our evaluation tasks is multilabel CAD classification, encompassing four cardiovascular disorders within ICD-10 categories I20–I25. Each disease is defined by the following codes:

\begin{itemize}[itemsep=4pt, topsep=2pt]
    \item Angina pectoris: I200, I201, I208, I209;
    \item Myocardial infarction: I210, I211, I212, I213, I214, I219;
    \item Other acute ischaemic heart diseases: I240, I248, I249;
    \item Chronic ischaemic heart disease: I250, I251, I252, I253, I254, I255, I256, I258, I259.
\end{itemize}

For CAD classification training and evaluation, we consider a subject positive only if they were diagnosed with CAD prior to the date of their cardiac MRI. Due to the nature of cardiovascular disease, it is possible that some subjects labeled as healthy may have undiagnosed CAD. Addressing this limitation in the UK Biobank is left for future work.

\section{Implementation Details}

\begin{table}[]
\centering
\caption{A list of tabular attributes and their UK Biobank field IDs that were used for the tabular similarity calculation at pretraining.}
\resizebox{\columnwidth}{!}{%
\begin{tabular}{ll}
\toprule
Tabular Feature & UK Biobank Field ID \\ \midrule
Sex & 31 \\
Smoking status & 20116 \\ 
Date of birth & 33 \\ 
Date I20 first reported (angina pectoris) & 131296 \\
Date of myocardial infarction & 42000 \\
Date I24 first reported (other acute ischaemic heart diseases) & 131304 \\
Date I25 first reported (chronic ischaemic heart disease) & 131306 \\
LVEF & Extracted \\
LVEDV & Extracted \\
LVESV & Extracted \\
LVSV & Extracted \\
LVEDM & Extracted \\
LVCO & Extracted \\
RVEDV & Extracted \\
RVESV & Extracted \\
RVSV & Extracted \\
RVEF & Extracted \\
RVCO & Extracted \\
MYOESV & Extracted \\
MYOEDV & Extracted \\
\bottomrule
\end{tabular}

}
\label{tab_attrlist}
\end{table}

\begin{table}[]
\centering
\caption{Tabular feature names and UK field IDs fo attributes that were not used for the pretraining, but included in the evaluations.}
\small
\begin{tabular}{ll}
\toprule
Tabular Feature & UK Biobank Field ID \\ \midrule
Height & 12144 \\
Weights & 21002 \\
Body mass index (BMI) & 21001 \\ 
\bottomrule
\end{tabular}

\label{tab_attrother}
\end{table}

\subsection{Baselines}

\begin{figure*}
    \centering
    \includegraphics[width=\linewidth]{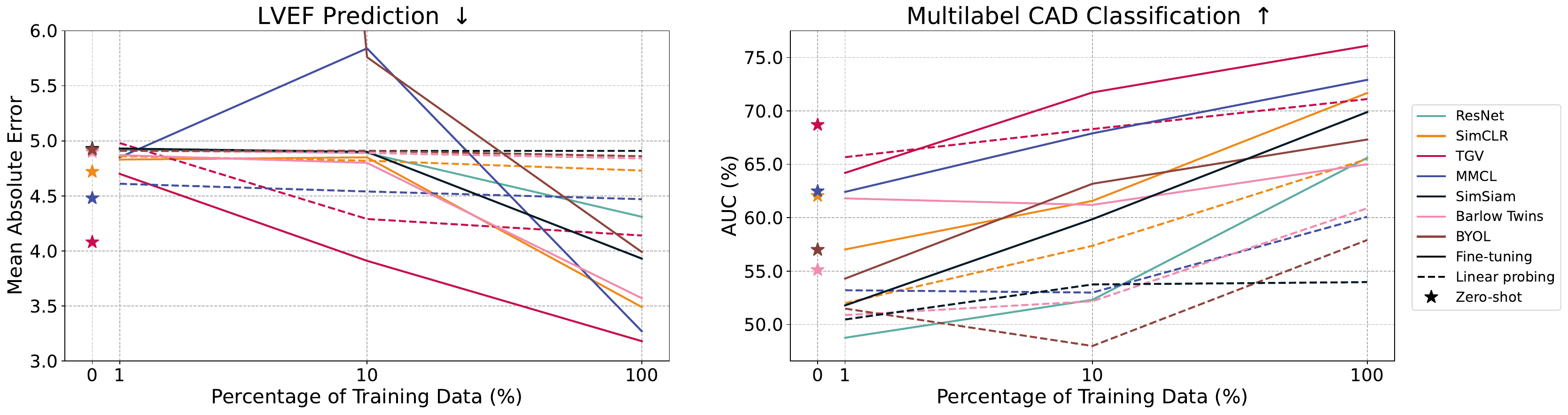}
    \caption{Performance of all evaluated methods on LVEF prediction and CAD classification using zero-shot prediction, and linear probing and fine-tuning under limited data regimes. The LVEF prediction result of BYOL \cite{grill2020bootstrap} with fine-tuning at 1\% is clipped; the value is 26.68. The reported results are obtained using strategies: fine-tuning (solid lines), linear probing (dashed lines), and zero-shot prediction (stars), applied consistently across all included methods.}
    \label{fig:limited_all}
\end{figure*}

\begin{table*}[]
    \centering
    \caption{The results of zero-shot predictions on TGV under varying size of representative set \(P\). The size is indicated by \(N\), with the performance of changes relative to  $N=2000$ shown in parenthesis. Red text indicated a decrease of performance. }
    \resizebox{\linewidth}{!}{%

\begin{tabular}{lccccccc}
\toprule
N    & CAD $\uparrow$ & LVEF $\downarrow$ & LVEDM $\downarrow$ & LVEDV $\downarrow$ & RVEF $\downarrow$ & RVEDV $\downarrow$ & MYOESV $\downarrow$  \\
\midrule
2000 &  68.70$\pm$0.83   &   4.08$\pm$0.00   &   7.64$\pm$0.04    &   13.63$\pm$0.07    &  3.98$\pm$0.01   & 12.43$\pm$0.05 & 8.18$\pm$0.06     \\
1000 &  67.83$\pm$1.43 \textcolor{red}{($\downarrow$-1\%)}   &   4.11$\pm$0.00 \textcolor{red}{($\uparrow$1\%)}   &   7.79$\pm$0.07 \textcolor{red}{($\uparrow$2\%)}    &   13.84$\pm$0.11 \textcolor{red}{($\uparrow$2\%)}  &  4.01$\pm$0.01  \textcolor{red}{($\uparrow$1\%)} & 12.63$\pm$0.03 \textcolor{red}{($\uparrow$2\%)}& 8.35$\pm$0.10 \textcolor{red}{($\uparrow$2\%)}     \\
500 &  67.58$\pm$1.22 \textcolor{red}{($\downarrow$-2\%)}  & 4.15$\pm$0.02 \textcolor{red}{($\uparrow$2\%)}  &   8.09$\pm$0.13 \textcolor{red}{($\uparrow$6\%)}   &   14.13$\pm$0.31 \textcolor{red}{($\uparrow$4\%)}    &  4.05$\pm$0.03 \textcolor{red}{($\uparrow$2\%)}   & 12.99$\pm$0.12 \textcolor{red}{($\uparrow$5\%)} & 8.62$\pm$0.13 \textcolor{red}{($\uparrow$5\%)}     \\
100 &  66.70$\pm$1.17 \textcolor{red}{($\downarrow$-3\%)}  & 4.32$\pm$0.04 \textcolor{red}{($\uparrow$6\%)}   &   8.74$\pm$0.45 \textcolor{red}{($\uparrow$14\%)}    &   15.34$\pm$0.61 \textcolor{red}{($\uparrow$13\%)}   &  4.30$\pm$0.06 \textcolor{red}{($\uparrow$8\%)}   & 14.34$\pm$0.55 \textcolor{red}{($\uparrow$15\%)} & 9.32$\pm$0.43  \textcolor{red}{($\uparrow$14\%)}    \\
\bottomrule
\end{tabular}

}
\label{tab_sets}
\end{table*}

We compare TGV against a mean-guess baseline (used only for cardiac phenotype prediction), a supervised 3D ResNet-50 model \cite{hara2018can}, four image-based contrastive learning approaches, and one image-tabular contrastive learning method. This section details the implementation of each baseline.

\vspace{4pt}
\noindent \textbf{Mean-guess. } The mean-guess baseline is applied to numerical cardiac phenotype values. The mean value for each phenotype is computed from the training set and used as the prediction on the test set. We report the mean absolute error (MAE) on the test set, calculated as the average difference between the ground truth and the calculated mean.

\vspace{4pt}
\noindent \textbf{Supervised. } We include one supervised baseline: 3D ResNet-50 \cite{hara2018can}. We use the implementation available in the MONAI framework \cite{cardoso2022monai}, training the model separately for each task using the same hyperparameters as described in Sec. \ref{sec_ft}.

\vspace{4pt}
\noindent \textbf{Contrastive. } All contrastive learning baselines are built on the 3D ResNet50 architecture \cite{hara2018can,cardoso2022monai}. SimCLR \cite{chen2020simple} is implemented using the NT-Xent loss provided by the lightly framework \cite{lightly}. The remaining contrastive methods, including image augmentation-based approaches such as BYOL \cite{grill2020bootstrap}, Barlow Twins \cite{zbontar2021barlow}, and SimSiam \cite{chen2021incremental}, as well as the image-tabular method MMCL \cite{hager2023best}, are based on a publicly available repository\footnote{\url{https://github.com/paulhager/MMCL-Tabular-Imaging}} implemented by Hager et al. \cite{hager2023best}. We adapt this codebase to support 3D ResNet-50 and modify it to match our 10-epoch training schedule, ensuring fair comparison. All contrastive baselines use the same image augmentation pipeline, designed to be compatible with 3D+t medical imaging:

\begin{itemize}
\vspace{2pt}
\item RandomHorizontalFlip(probability=0.5),
\vspace{4pt}
\item RandomResizedCrop(size=128, scale=(0.6, 1.0)),
\vspace{4pt}
\item RandomRotation(degrees=45).
\vspace{2pt}
\end{itemize}

All augmentations are implemented using the torchvision library and were selected for their suitability to medical data and volumetric inputs.

\begin{table*}[]
    \centering
     \caption{Mean and standard deviation of the zero-shot prediction over three non-overlapping representative sets \(P\). The best results is \textbf{bold}, while the second best is \underline{underlined}. }
    \resizebox{\linewidth}{!}{%
\small
\begin{tabular}{lccccccccccc}
\toprule
 Model    & CAD \(\uparrow\) & LVEF \(\downarrow\) & LVEDM \(\downarrow\) & LVEDV \(\downarrow\) & RVEF \(\downarrow\) & RVEDV \(\downarrow\) & MYOESV \(\downarrow\) & BMI \(\downarrow\) & Height \(\downarrow\) & Weight \(\downarrow\) & Sex \(\uparrow\)\\
     \midrule
\textit{Image Augmentation} &  &  &  &  &  &  &  & & & & \\
SimCLR \cite{chen2020simple} & 62.05$\pm$1.36 & 4.72$\pm$0.00 & 13.01$\pm$0.07 & 23.26$\pm$0.11 & 4.71$\pm$0.02 & 21.89$\pm$0.16  & 13.48$\pm$0.41  & \underline{2.54$\pm$0.03} & 6.24$\pm$0.03 & \underline{8.43$\pm$0.08} & 75.04$\pm$0.66 \\
BYOL \cite{grill2020bootstrap} & 56.99$\pm$0.79 & 4.92$\pm$0.01  & 16.43$\pm$0.01  & 27.81$\pm$0.19 & 4.95$\pm$0.03 & 26.29$\pm$0.19 & 16.19$\pm$0.37 & 3.33$\pm$0.44 & 7.02$\pm$0.01 & 10.44$\pm$0.13 & 69.72$\pm$0.36    \\
SimSiam \cite{chen2021exploring}  & 57.01$\pm$1.58 & 4.93$\pm$0.01 & 16.04$\pm$0.12 & 27.46$\pm$0.09 & 4.94$\pm$0.04 & 25.62$\pm$0.22 & 16.14$\pm$0.13 & 2.86$\pm$0.02 & 6.93$\pm$0.03 & 8.80$\pm$1.30 & 68.19$\pm$0.31 \\
Barlow Twins \cite{zbontar2021barlow}  & 55.12$\pm$1.59 & 4.90$\pm$0.04 & 16.54$\pm$0.21 & 27.17$\pm$0.04 & 4.97$\pm$0.08 & 25.98$\pm$0.14 & 16.52$\pm$0.17 & 3.16$\pm$0.03 & 7.02$\pm$0.06 & 10.88$\pm$0.09 & 72.90$\pm$0.35  \\ \midrule
\textit{Tabular Supervision} &  &  &  &  &  &  &  &  & & \\ 
MMCL \cite{hager2023best}  & \underline{62.49$\pm$0.50} & \underline{4.48$\pm$0.06} & \underline{8.74$\pm$0.11}& \underline{15.12$\pm$0.07} & \underline{4.55$\pm$0.05} & \underline{13.70$\pm$0.12} & \underline{9.54$\pm$0.13} & 2.96$\pm$0.07 & \underline{5.13$\pm$0.03} & 8.68$\pm$0.24  & \underline{92.60$\pm$0.15} \\ \midrule
\textit{Tabular Guidance} &  &  &  &  &  &  &  & & & &   \\
TGV (Ours) & \textbf{68.70$\pm$0.83} & \textbf{4.08$\pm$0.00} & \textbf{7.64$\pm$0.04} & \textbf{13.63$\pm$0.07} & \textbf{3.98$\pm$0.01} & \textbf{12.43$\pm$0.05}  & \textbf{8.18$\pm$0.06}  & \textbf{2.39$\pm$0.08} & \textbf{4.88$\pm$0.02} & \textbf{7.44$\pm$0.07} & \textbf{98.16$\pm$0.13} \\
\bottomrule
\end{tabular}
}
\label{tab_std}
\end{table*}

\begin{figure*}
    \centering
    \includegraphics[width=\linewidth]{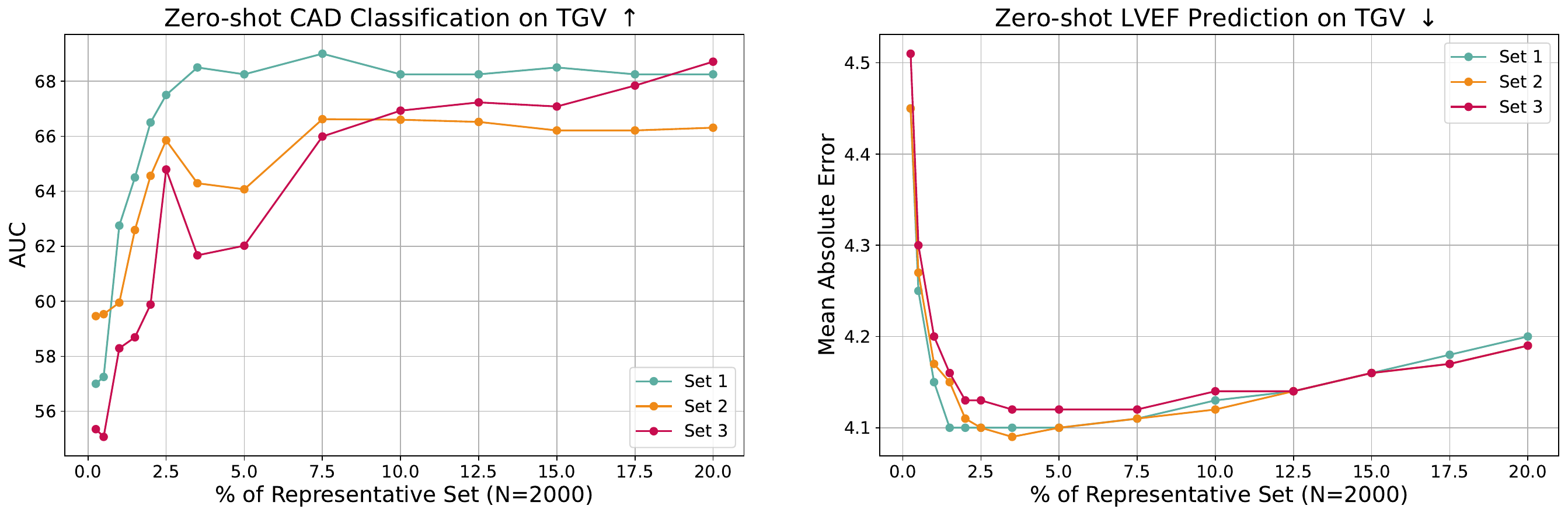}
    \caption{Zero-shot CAD classification and LVEF prediction performance as a function of the percentage of samples used for the mean aggregation out of all the samples in the representative set, i.e., number of nearest neighbors used for making the prediction (MP Eq. 9). The size of the representative sets is N=2,000. Results are reported on the validation set; the best-performing configuration is selected for final evaluation on the test set.}
    \label{fig:zs_empirical}
\end{figure*}

\subsection{Fine-tuning \& Linear Probing} \label{sec_ft}

We use the same training hyperparameters for both fine-tuning and linear probing. All downstream tasks are trained for 35 epochs. Cardiac phenotype prediction and CAD classification use a learning rate of 1e-3 and 1e-4 respectively. Cardiac phenotype prediction is trained without augmentations under the full data regime, and with an augmentation rate of 0.6 in the 10\% and 1\% data settings. CAD classification uses a fixed augmentation rate of 0.4 across all data regimes. We employ L1 loss for cardiac phenotype prediction and binary cross-entropy with logits for CAD classification.

\subsection{Zero-shot Prediction}

Our zero-shot prediction approach is based on using a representative sample of images to compare the query image against. These representative subsamples are drawn from the disease-balanced training set used for fine-tuning the CAD classification model. The size of the disease balanced data (6,424 samples) allows to construct three non-overlapping subsamples, each containing 2,000 images. The final result is reported by averaging the predictions across these three subsamples. 

\section{Additional Cardiac Experiments}

\begin{table*}[]
\centering
\begin{tabular}{c}
    \begin{minipage}{\linewidth}
        \centering
        \includegraphics[width=\linewidth]{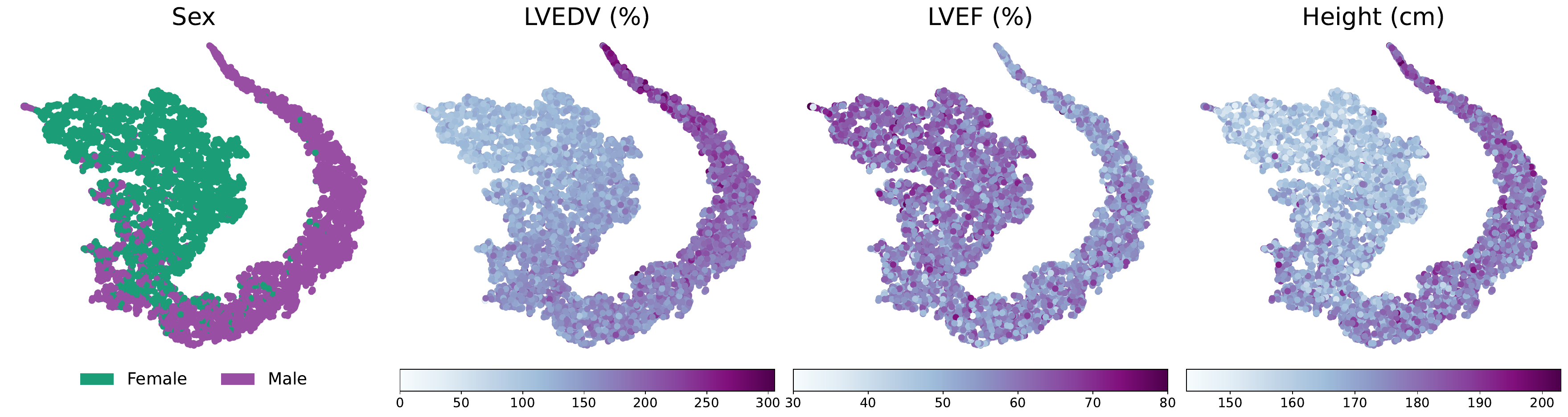}
        \small (a) MMCL
    \end{minipage}
    \\
    \begin{minipage}{\linewidth}
        \centering
        \includegraphics[width=\linewidth]{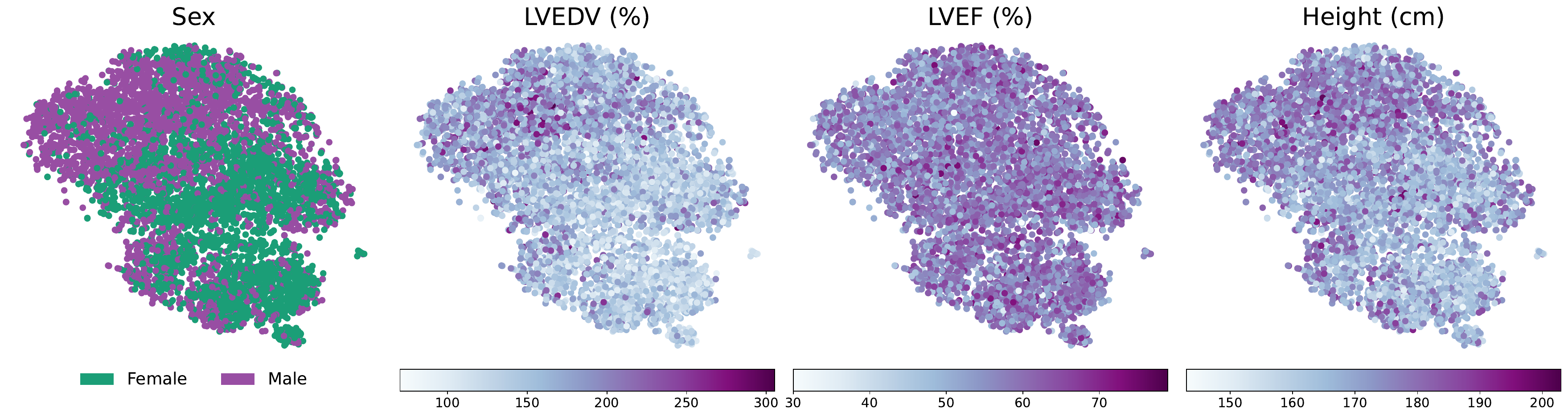}
        \small (b) SimCLR
    \end{minipage}
\end{tabular}
\captionof{figure}{t-SNE visualization of sex, LVEDV, LVEF, and height for a) MMCL \cite{hager2023best} and b) SimCLR \cite{chen2020simple}. Sex, LVEDV, and LVEF were included as attributes in the pretraining of MMCL, height was not. SimCLR is an image augmentation based method. Thus, no tabular attributes were used during its training. }
\label{fig_tsneother}
\end{table*}

\subsection{Performance under Low-Data Regimes (Complete)}

Fig. \ref{fig:limited_all} presents the results on CAD classification and LVEF prediction under low-data regimes for all the baselines, which were omitted for clarity in the main body of the paper. TGV outperforms the other methods on nearly all the data regimes and all tasks, with some exceptions. MMCL \cite{hager2023best} and SimCLR \cite{chen2020simple} are typically the second best approach, while BYOL \cite{grill2020bootstrap}, Barlow Twins \cite{zbontar2021barlow}, and SimSiam \cite{chen2021exploring} report the worst overall performance. 

\subsection{Evaluating Robustness of the Zero-shot Predictions}

We evaluate the robustness of our zero-shot approach in terms of two conditions: (1) how changing the size of the representative set impacts performance, and (2) how changing the samples included in the representative set affects the predictions.

\vspace{4pt}
\noindent \textbf{1) Robustness to representative set size. } We evaluate the robustness of the zero-shot predictions under different sizes of the representative set \(P\). The experiment is performed using the image encoder pretrained with TGV and the results are reported in Table \ref{tab_sets}. We consider the N=2000 as the baseline and report the changes in the performance against it. Reducing the size of the representative set leads to a worsening in the performance of the zero-shot prediction, as some target values might end up being underrepresented. However, the performance even at N=100, still outperforms the results of mean-guess for cardiac phenotype prediction and is comparable with supervised ResNet for both CAD classification and cardiac phenotype prediction (Main Paper (MP) Table 1), showing the robustness of our zero-shot prediction approach. 

\vspace{4pt}
\noindent \textbf{2) Robustness across different representative sets. } Table \ref{tab_std} reports the mean and standard deviation of zero-shot prediction performance across three different representative sets \(P\). CAD prediction shows the highest standard deviation, which is reflective of the small number of CAD positive cases in the UK Biobank. Generally, methods with lower overall performance also exhibit higher standard deviation, suggesting that stronger representations yield more robust zero-shot predictions that are less sensitive to changes in the representative sets.

\subsection{Nearest Neighbor Number Selection for Zero-Shot Prediction} 
The final zero-shot prediction is computed as the mean of the K nearest neighbors of the query image (MP Eq. 9). The number K is chosen empirically based on validation set performance. Fig. \ref{fig:zs_empirical} illustrates the results across different values of K, expressed as a percentage of the representative set P, i.e., the amount of samples from P used for mean aggregation, for both CAD classification and LVEF prediction. The optimal K varies depending on the task and representative set. Choosing too few neighbors may yield unrepresentative predictions, while too many can bias results toward simply outputting the mean value of the representative set, especially for cardiac phenotype prediction. Therefore, selecting an appropriate K is crucial to maximize performance for the given task and data.

\subsection{Representation Evaluation using t-SNE}
We evaluate the representations generated by MMCL \cite{zhang2022contrastive} and SimCLR \cite{chen2020simple} using t-SNE. Fig. \ref{fig_tsneother} presents visualizations of the representations with respect to sex, LVEDV, LVEF, and height. Among these four attributes, only height was not included in the pretraining of MMCL. SimCLR, being an image augmentation-based method, did not use any tabular data during pretraining. MMCL demonstrates superior clustering ability compared to SimCLR across all attributes. While the comparison between SimCLR and MMCL already underscores the importance of tabular data-informed pretraining, TGV produces even more coherent and clinically meaningful clusters (MP Fig. 4), further solidifying the value of incorporating tabular data into contrastive learning frameworks.

\section{Assessing TGV's Generalizability}

\subsection{Dataset}

To assess whether TGV can generalize to other domains and datasets, we use the Data Visual Marketing (DVM) car dataset \cite{huang2023dvmcarlargescaleautomotivedataset}. The dataset contains 1,451,784 images and their corresponding attributes of cars at varying degree angles. Model performance is evaluated on two tasks, car model classification (286 classes) and price prediction (regression). We use the same preprocessing technique as MMCL \cite{hager2023best}, which yields 70,565 image-tabular data training pairs, 17,642 validation pairs, and 88,207 test pairs. The tabular data attributes include information regarding width, length, height, wheelbase, price, advertisement year, miles driven, number of seats, number of doors, original price, engine size, body type, gearbox type, fuel type, color, and car model. MMCL achieved the best results while using all of the listed attributes, while TGV performed best without including the advertisement year, miles driven, and color. Similarly to our observations on the cardiac dataset, the choice of attributes for TGV is critical; including unrelated attributes leads to noisy and less meaningful similarity scores. For both methods, we report results using their respective best sets of tabular attributes. 

\begin{table}[]
    \centering
        \caption{The results on car model prediction (accuracy $\uparrow$) and price prediction (MAE $\downarrow$) on the DVM car dataset. ZS stands for zero-shot, LP for linear probing, and FT for fine-tuning. }
    \resizebox{\columnwidth}{!}{%
        \begin{tabular}{lcccccc}
\toprule

           & \multicolumn{3}{c}{Model Classification \(\uparrow\)} & \multicolumn{3}{c}{Price Regression \(\downarrow\)} \\
\midrule
           
           & ZS      & LP     & FT     & ZS      & LP     & FT     \\ 
\midrule
           
Mean-Guess & -       & -      & -      &     8752.2    &   8752.2    &     8752.2  \\
ResNet50 \cite{he2016deep}    & -       &    -    &    92.95    & -       &   -     &    3411.4    \\
SimCLR \cite{chen2020simple}     &  13.52       &    56.01    &    89.53    &   6609.5      &   5701.3     &   3674.7     \\
MMCL  \cite{hager2023best}     &    81.06     &    91.64    &    94.06    &     3621.1    &   4440.6     &   2821.6     \\
TGV (Ours) &   \textbf{83.37}      &    \textbf{92.52}    &    \textbf{94.23}    &    \textbf{3529.6}     & \textbf{3759.6}       & \textbf{2650.6} \\
\bottomrule
\end{tabular}
    }
    \label{tab_demo}
\end{table}

\subsection{Implementation Details}

All the models use a 2D ResNet-50 as a backbone and are pretrained for 500 epochs under a batch size of 512. The best model is selected based on the best validation performance. For the downstream tasks, the models are further trained for 500 epochs for both linear probing and fine-tuning. This follows the setting of previous works \cite{hager2023best}. For the baseline models we use the hyperparameters as described by the authors. TGV is pretrained using a learning rate of 1e-3. Model and price predictions are both trained using a learning rate of 3e-4 and a batch size of 512. We use the best performing baselines on UK Biobank as our baselines for DVM, namely SimCLR \cite{chen2020simple} and MMCL \cite{hager2023best}. All the training data is used for the downstream task tuning of the methods, while the representative set is assembled using 10\% of the training data. 

\subsection{Does TGV work on natural images?}

We evaluate the generalizability of TGV to natural images by employing the DVM car dataset and evaluation the trained representations on two tasks, car model prediction and price regression. The car model prediction is evaluated using accuracy and price prediction using mean absolute error (MAE). TGV achieves the best performance on both car model classification and price regression using every tuning scheme. This shows that our approach is capable of generalizing to 2D data and natural images without incorporating any changes to the method. Both UK Biobank \cite{sudlow2015uk} and DVM datasets are relatively homogeneous, each representing a single organ (heart) or item (car). While the images are often visually similar, tabular data–based similarity provides guidance signals that can capture even subtle differences and enable stronger representations for downstream tasks.

\end{document}